\documentclass{article}
\usepackage{spconf,amsmath,graphicx}

\usepackage{enumitem}
\setlist{nosep, leftmargin=14pt}
\usepackage{xcolor}
\usepackage{booktabs,multirow}
\usepackage{amssymb}
\usepackage{hyperref}

\title{An Automated Framework for Large-Scale Graph-Based Cerebrovascular Analysis}
\name{
\begin{tabular}{@{}c@{}}
Daniele Falcetta$^{1}$, Liane S. Canas$^{2}$, Lorenzo Suppa$^{1,3}$,  Matteo Pentassuglia$^{1}$, \\
\textit{Jon Cleary}$^{2}$, \textit{Marc Modat}$^{2}$, \textit{S\'{e}bastien Ourselin}$^{2}$, \textit{Maria A. Zuluaga}$^{1,2}$
\end{tabular}
}

\address{
$^{1}$ EURECOM, Sophia Antipolis, France \\
$^{2}$ School of Biomedical Engineering \& Imaging Sciences, King's College London, UK \\
$^{3}$ Politecnico di Torino, Torino, Italy
}
\begin{document}

\maketitle
\begin{abstract}
We present \verb|CaravelMetrics|, a computational framework for automated cerebrovascular analysis that models vessel morphology through skeletonization-derived graph representations. The framework integrates atlas-based regional parcellation, centerline extraction, and graph construction to compute fifteen morphometric, topological, fractal, and geometric features. The features can be estimated globally from the complete vascular network or regionally within arterial territories, enabling multiscale characterization of cerebrovascular organization. Applied to 570 3D TOF-MRA scans from the IXI dataset (ages 20--86), \verb|CaravelMetrics| yields reproducible vessel graphs capturing age- and sex-related variations and education-associated increases in vascular complexity, consistent with findings reported in the literature. The framework provides a scalable and fully automated approach for quantitative cerebrovascular feature extraction, supporting normative modeling and population-level studies of vascular health and aging.
\end{abstract}


\section{Introduction}
\label{sec:intro}

Brain blood vessels change throughout the lifespan, with vessel density declining, curvature increasing, and branching patterns reorganizing, with implications for brain health~\cite{Bennett2024aging}. Understanding cerebrovascular morphology is critical for detecting disease and distinguishing pathological from normal changes.
To that end, population-level analyses are crucial as individual vessel measurements cannot be interpreted without demographic-specific reference ranges. These reference ranges allow discrimination between healthy aging and pathological changes across groups, and identification of clinical biomarkers that distinguish disease-related alterations from normal variations within the population. However, large-scale analyses have been limited by manual vessel tracing and a lack of automated biomarker extraction tools that can be used at scale.

Current approaches for cerebrovascular analysis can be classified into three categories. 
Segmentation-based methods, enabled by recent advances in deep learning for accurate automated vessel segmentation, typically from time-of-flight magnetic resonance angiography (TOF-MRA), estimate arterial volume from the segmented masks at the whole-brain and regional levels~\cite{Guo2024cerevesseg}.
Atlas-based approaches provide population references through statistical arterial atlases~\cite{Mouches2019atlas}, offering normative cerebrovascular anatomy baselines. However, both approaches focus on simple volumetric and diameter measurements that may fail to completely characterize the vascular tree.
Finally, graph-based computational tools use a segmentation mask as input to generate a graph representation of the vascular tree, from which intensity and morphometric features are extracted~\cite{chen2018development,Spangenberg2023vesselexpress,Bumgarner2022vesselvio,Li2025lumen,yang2025rvpd}.
While commonly used, these tools are either semi-automatic~\cite{chen2018development,Li2025lumen} or have been specifically designed for a particular type of image~\cite{Spangenberg2023vesselexpress,Bumgarner2022vesselvio,yang2025rvpd}, limiting their usage at scale for any neurovascular imaging modality.

\begin{figure*}[t]
    \centering
    \includegraphics[width=0.85\textwidth]{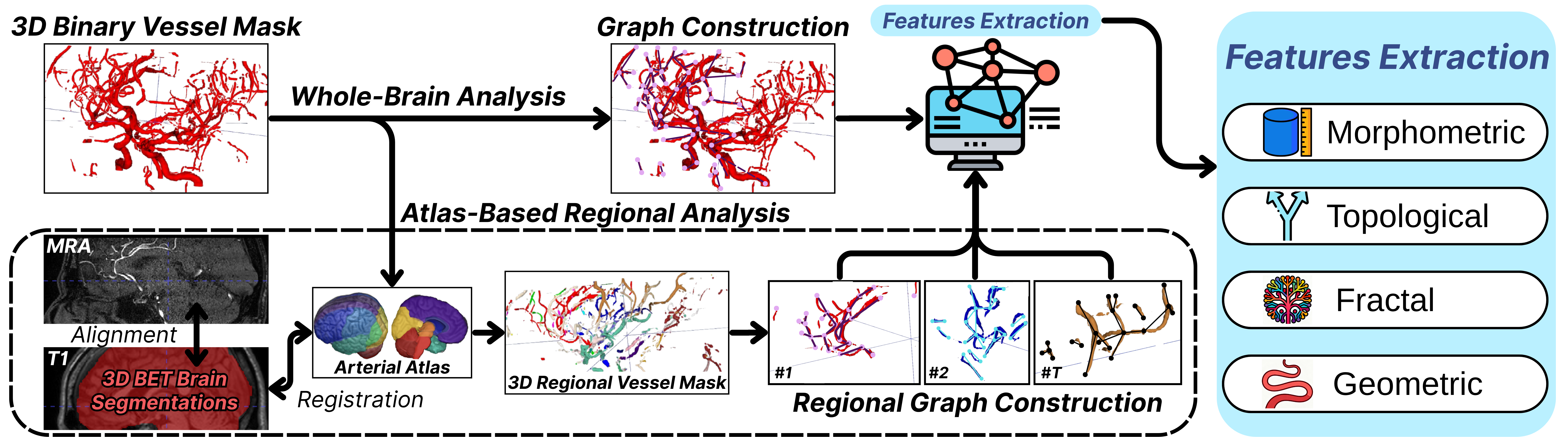}
    \caption{Overview of the vascular feature extraction framework.}
    \label{fig:pipeline}
\end{figure*}

\begin{table*}[t]
\centering
\caption{List of vascular features extracted within the vessel analysis framework.}
\footnotesize
\begin{tabular}{p{2.2cm}p{3.5cm}p{6.5cm}p{3cm}}
\toprule
\textbf{Category} & \textbf{Metric} & \textbf{Definition/Formulation} & \textbf{Application} \\
\midrule
\multirow{2}{2.2cm}{\centering\textbf{Morphometric}} 
& Total vessel length & $L = \sum_{(u,v) \in E} d_{uv}$ & Structural integrity \\
& Volume & $|M| \, \Delta x \, \Delta y \, \Delta z$ & Vascular volume \\
\midrule
\multirow{6}{2.2cm}{\centering\textbf{Topological}}
& Bifurcation count & $N_{bif} = |\{v \in V : \deg(v) = 3\}|$ & \multirow{2}{3cm}{Branching complexity} \\
& Bifurcation density & $N_{bif}/L$ & \\
\cmidrule(lr){2-3}
& Number of loops & $|E_k| -|V_k| + 1$ & \multirow{2}{3cm}{Connectivity patterns} \\
& Loop lengths & $\sum_{(u,v) \in \text{loop}} d_{uv}$ for each loop & \\
\cmidrule(lr){2-3}
& Abnormal-degree nodes & $|\{v \in V : \deg(v) > 3\}|$ & Anomalous branching \\
& Connected components & $|\{G_k\}|$ 
& Network integrity \\
\midrule
\multirow{2}{2.2cm}{\centering\textbf{Fractal}}
& Fractal dimension & $-\frac{\Delta \log(N(\varepsilon))}{\Delta \log(\varepsilon)};$ \, $N(\varepsilon)$: count of boxes at scale $\varepsilon$ & Multi-scale organization \\
& Lacunarity & $\frac{\text{Var}(n)}{[\text{Mean}(n)]^2} + 1$; \, $n$ the number of voxels per box & Spatial heterogeneity \\
\midrule
\multirow{5}{2.2cm}{\centering\textbf{Geometric}}
& Geodesic length & $L_{geo}(P) = \sum_{i=1}^{n-1} d_{v_i v_{i+1}}$  
& Segment arc length \\
& Instantaneous curvature & $\kappa(t) = \frac{\|\mathbf{x}'(t) \times \mathbf{x}''(t)\|}{\|\mathbf{x}'(t)\|^3}$ on interpolated $\mathbf{x}(t)$ & Local geometry \\
& Mean curvature & $\bar{\kappa} = \frac{1}{L_{geo}} \int_0^1 \frac{\kappa(t)}{n(t)} \|\mathbf{x}'(t)\| dt$ & Average tortuosity \\
& Mean squared curvature & $\frac{1}{L_{geo}} \int_0^1 \frac{\kappa^2(t)}{[n(t)]^2} \|\mathbf{x}'(t)\| dt$ & Curvature variability \\
& Arc-over-chord ratio & $L_{geo}(P) / \|\mathbf{x}_{v_1} - \mathbf{x}_{v_n}\|$ & Segment tortuosity \\
\bottomrule
\end{tabular}

\label{tab:features}
\end{table*}

In this work, we present \verb|CaravelMetrics|, an automated framework for analyzing the cerebrovascular tree. Our framework extracts fifteen morphometric, topological, fractal, and geometric vascular features from a graph generated from vascular binary masks, thus designed to be agnostic to the underlying image modality. Similar to~\cite{Guo2024cerevesseg}, we integrate an atlas of arterial regions, allowing us to analyze the cerebrovascular tree at both the whole-brain and regional levels. We showcase the potential of our open-source tool by analyzing the IXI Dataset\footnote{\href{https://brain-development.org/ixi-dataset/}{https://brain-development.org/ixi-dataset/}}, a cohort of 570 subjects aged 20 to 86 years whose cerebrovascular trees were imaged using TOF-MRA. Our analysis reveals that our features effectively capture variations in the vessels across a population.

\section{Methods}
\verb|CaravelMetrics| relies on representing the cerebrovascular network as a mathematical graph structure that is extracted from a vascular binary mask. The graph and the vascular binary mask are then used to extract morphometric, topological, fractal, and geometric features that can be aggregated at the whole-brain or regional level using an arterial atlas (Figure \ref{fig:pipeline}). 

\vspace{4pt}
\noindent\textbf{Graph Representation of Cerebrovascular Networks.}
Let us denote $M: \Omega \subset \mathbb{Z}^3 \to \lbrace0,1\rbrace$, a binary vascular mask obtained from any 3D imaging modality. Given $M$, we define its baseline skeleton $S = \operatorname{Skel}(M) \subseteq \Omega$ as a one-voxel-thick subset that preserves the topological properties of $M$, such that $S \subseteq M$, $\operatorname{conn}(S) = \operatorname{conn}(M)$, and $\operatorname{hom}(S) = \operatorname{hom}(M)$, where $\operatorname{conn}(\cdot)$ denotes the number of connected components and $\operatorname{hom}(\cdot)$ describes the topology (loops and holes).
From the skeleton points $\mathbf{x} = (x, y, z) \in S \subset \mathbb{Z}^3$, we construct an undirected weighted graph $G = (V, E)$, where
each skeleton point corresponds to a node $v \in V$ with spatial coordinates $\mathbf{x}_v \in \mathbb{Z}^3$. Edges $E = \{ (u, v) \}$ connect pairs of nodes within a $3 \times 3 \times 3$ neighborhood (26-connectivity).

After graph construction, two post-processing steps are applied. First, spurious edges are removed by sampling intensity along each edge and retaining low-intensity edges if $d_{geo} < 2 d_{euc}$, with $d_{geo}$ and $d_{euc}$ the geodesic and Euclidean distances between two edge endpoints. Second, to maximize coverage, \textit{orphan} points, $\mathbf{x} \in M \setminus S$ with $\min_{s\in S}d_{uv} < D$, are identified and integrated via centroid connections under geodesic and branching-angle constraints.

A vascular graph $G$ may consist of $k$ disconnected subgraphs, each corresponding to an individual vessel component, such that $G = \bigsqcup_k G_k$.
Within a given $G_k$, a root node serves as a reference point for segment extraction. We define three root nodes to ensure representative coverage of the vascular structure: $R_1$ corresponds to the end-point ($\deg(v) = 1$) with the largest vessel diameter, $R_2$ to the end-point with the second-largest diameter, and $R_3$ to the bifurcation node ($\deg(v) \geq 3$) with the largest diameter. If no bifurcation exists, $R_3$ defaults to $R_1$. Based on this, a \textit{vessel branch} or \textit{segment} denotes the shortest path between a selected root node $s \in \{R_1, R_2, R_3\}$ and each reachable end-point $e \in V_{\text{end}}$ within the same $G_k$:
\begin{equation}
\label{eq:path}
   P_{s \to e} = \arg\min_{p \in \mathcal{P}(s,e)} \sum_{(u,v) \in p} d_{uv}, 
\end{equation}
where $\mathcal{P}(s,e)$ denotes the set of paths between $s$ and $e$ in $G_k$.

%\vspace{3pt}
\noindent\textbf{Feature Extraction.}
The framework extracts 15 vascular features, grouped into 4 categories (Table~\ref{tab:features}), each capturing a distinct property of cerebrovascular organization. 
Morphometric features quantify network extent and capacity through total vessel length and volume. 
Topological features describe branching architecture and connectivity using bifurcation count, loop number, and component analysis. 
Fractal features characterize hierarchical self-similarity via box-counting and lacunarity~\cite{Dougherty2011multifractal}.
Finally, geometric features quantify vessel tortuosity, reflecting arterial stiffening and age-related alterations, by capturing local vessel morphology and shape irregularity through cubic-spline parameterization of discrete centerline segments (Eq.~\ref{eq:path}) as continuous curves \(\mathbf{x}(t):[0,1]\!\to\!\mathbb{R}^3\), with curvature values weighted by path multiplicity \(n(t)\) and averaged across root nodes \(s\in\{R_1,R_2,R_3\}\) to obtain descriptors for each $G_k$.

\vspace{3pt}
\noindent\textbf{Atlas-Based Regional Analysis.}
\begin{figure}
    \centering
    \includegraphics[width=0.9\linewidth]{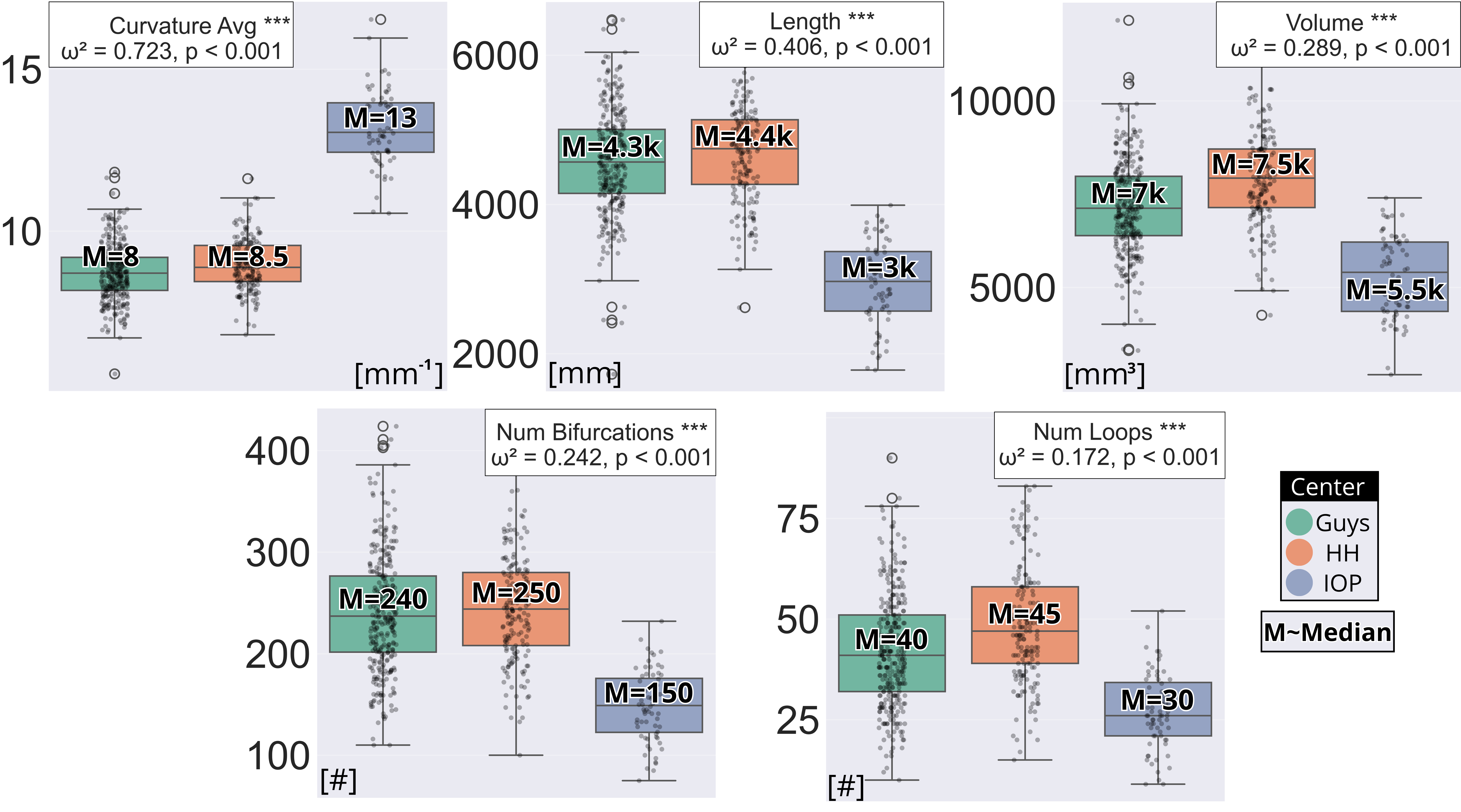}
    \caption{\textbf{Scanner-induced systematic bias.} Consistent IOP shift across all features demonstrates acquisition artifacts.}
    \label{fig:site_effects}
\end{figure}
The cerebrovascular tree exhibits heterogeneous patterns across brain regions, motivating a territory-based analysis. To enable this, an arterial atlas $\mathcal{A} = \lbrace T_i \rbrace_{i=1}^T$, which partitions the brain into $T$ arterial territories, is aligned to the neurovascular image, yielding territory-specific masks $M_i = M \cap T_i$.
Our framework supports two complementary analysis modes: \textit{whole-brain analysis}, on the full graph \( G = (V, E) \) derived from $M$; and \textit{region-specific analysis}, on regional graphs $G_i = (V_i, E_i)$ derived from $M_i$. 
This formulation enables global or arterial-territory-specific feature extraction.

\vspace{3pt} 

\noindent\textbf{Implementation Details}. Skeletonization, graph generation, and feature extraction are implemented in Python 3.10 with NumPy, SciPy, scikit-image, vedo, and NetworkX. Orphan connections are validated by a $30^\circ$ branching-angle constraint and a distance constraint $D>2.5$ mm. Graph geometry is refined via iterative Laplacian smoothing ($N_{iter}=2, \alpha=0.8$) and components with $<$ 7 nodes are removed. We employ the open-source 3D brain MRI arterial territories atlas proposed in~\cite{liu2023digital} ($T=30$). Atlas-to-image alignment is done through multi-step registration: brain extraction (FSL BET \cite{Jenkinson2012fsl}), reorientation to MNI space, atlas-to-T1w registration using FLIRT with mutual information, and T1w-to-TOF-MRA co-registration.
Code and documentation are available in the project's repository\footnote{\href{https://github.com/i-vesseg/CaravelMetrics}{https://github.com/i-vesseg/CaravelMetrics}}.

\begin{figure}
    \centering
    \includegraphics[width=0.75\linewidth]{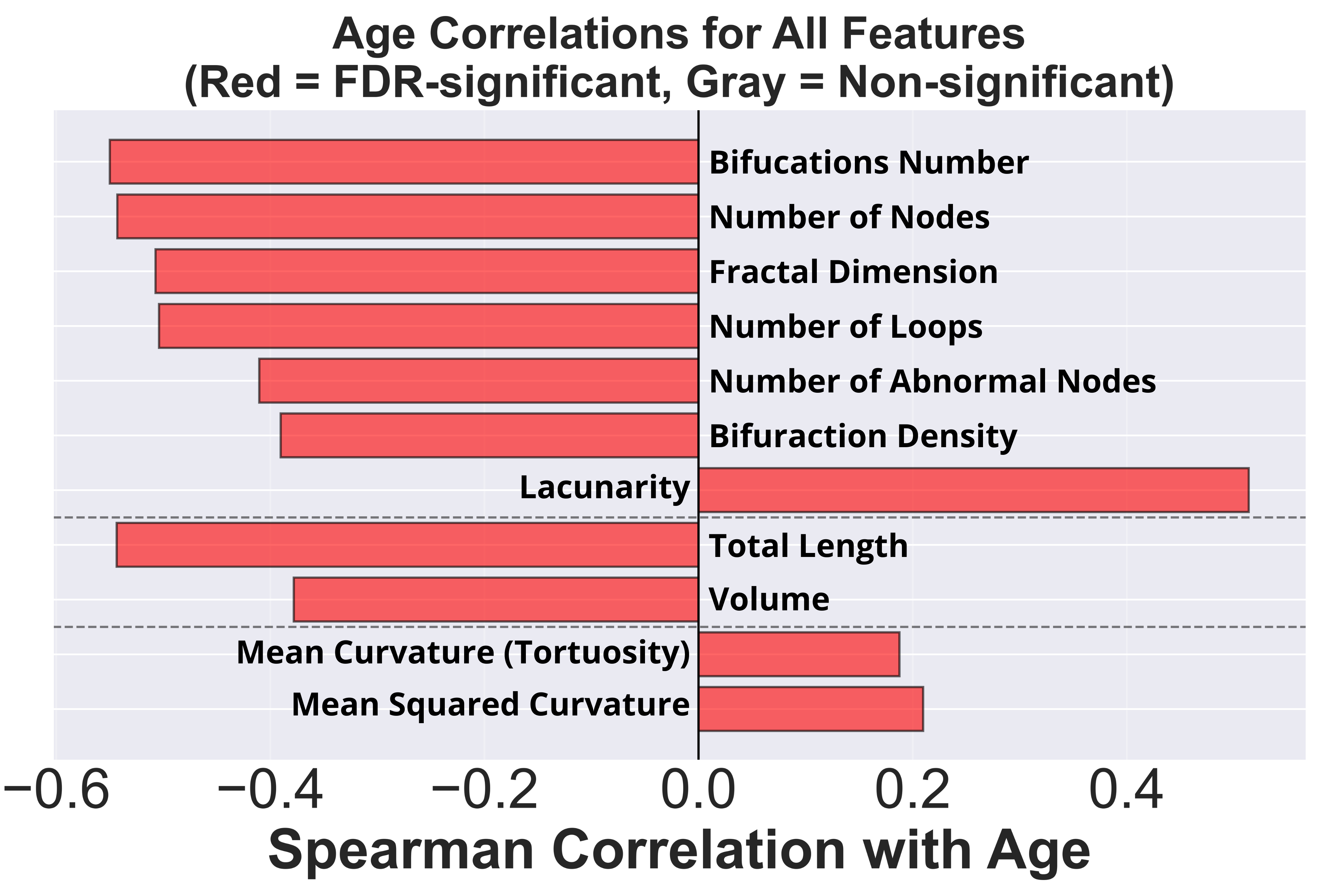}
    \caption{\textbf{Age effects on vessel morphometry.} Significant age correlations (red bars) after FDR correction.} 
    \label{fig:age_effects}
\end{figure}
\section{Experiments and Results}
\label{sec:exps}
We evaluate the proposed framework by analyzing 570 healthy subjects (20--86 years) from the IXI dataset, with multiple MR sequences acquired across three London hospitals (Guys, HH, and IOP) with different scanners. Each subject has associated demographic information, including age, sex, height, weight, education, and occupation.Body mass index (BMI) is computed from height and weight, and subjects are stratified into 4 standard BMI categories.

Using IXI's TOF-MRA deep learning-generated binary vessel masks from the VesselVerse dataset~\cite{FalDan_VesselVerse_MICCAI2025}, we apply \verb|CaravelMetrics| to quantify age-related cerebrovascular changes and demographic associations with vessel morphology, tortuosity, and network topology, consistent with the literature (e.g.,~\cite{Bullitt2010aging,ophelders2024anatomical}). Atlas-based regional analysis uses T1w IXI images for multi-step registration. Age effects are assessed using Spearman correlations ($r$) and one-way ANOVA across quartile-based groups with $\eta^2$ effect sizes. Demographic comparisons utilize t-tests for binary factors (sex) and one-way ANOVA for categorical variables (BMI, education), with $\omega^2$ effect sizes for unbalanced groups. Multi-center effects undergo site-stratified analysis to separate biological variation from acquisition artifacts. All analyses apply Benjamini-Hochberg False Discovery Rate (FDR) correction ($q < 0.05$) to control for multiple comparisons.

\begin{figure}
    \centering
    \includegraphics[width=0.95\linewidth]{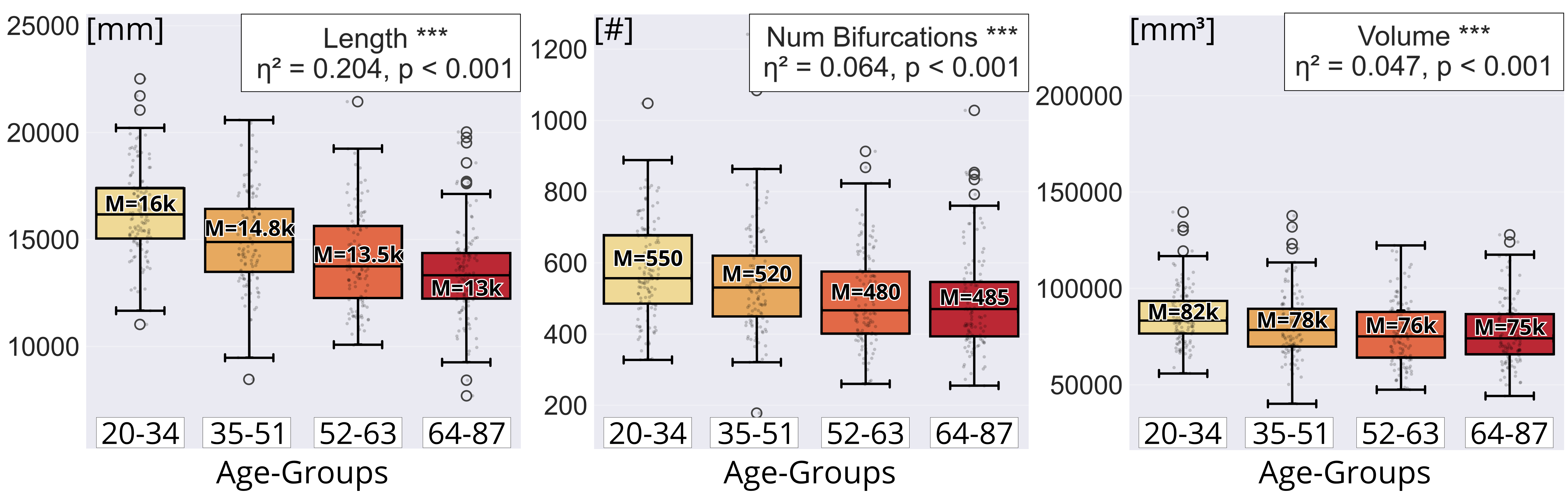}
    \caption{\textbf{Age-related vessel changes.} Group differences across multiple morphometric and topological features.}
\label{fig:age_site_comparison}
\end{figure}

\begin{figure}
    \centering
    \includegraphics[width=0.80\linewidth]{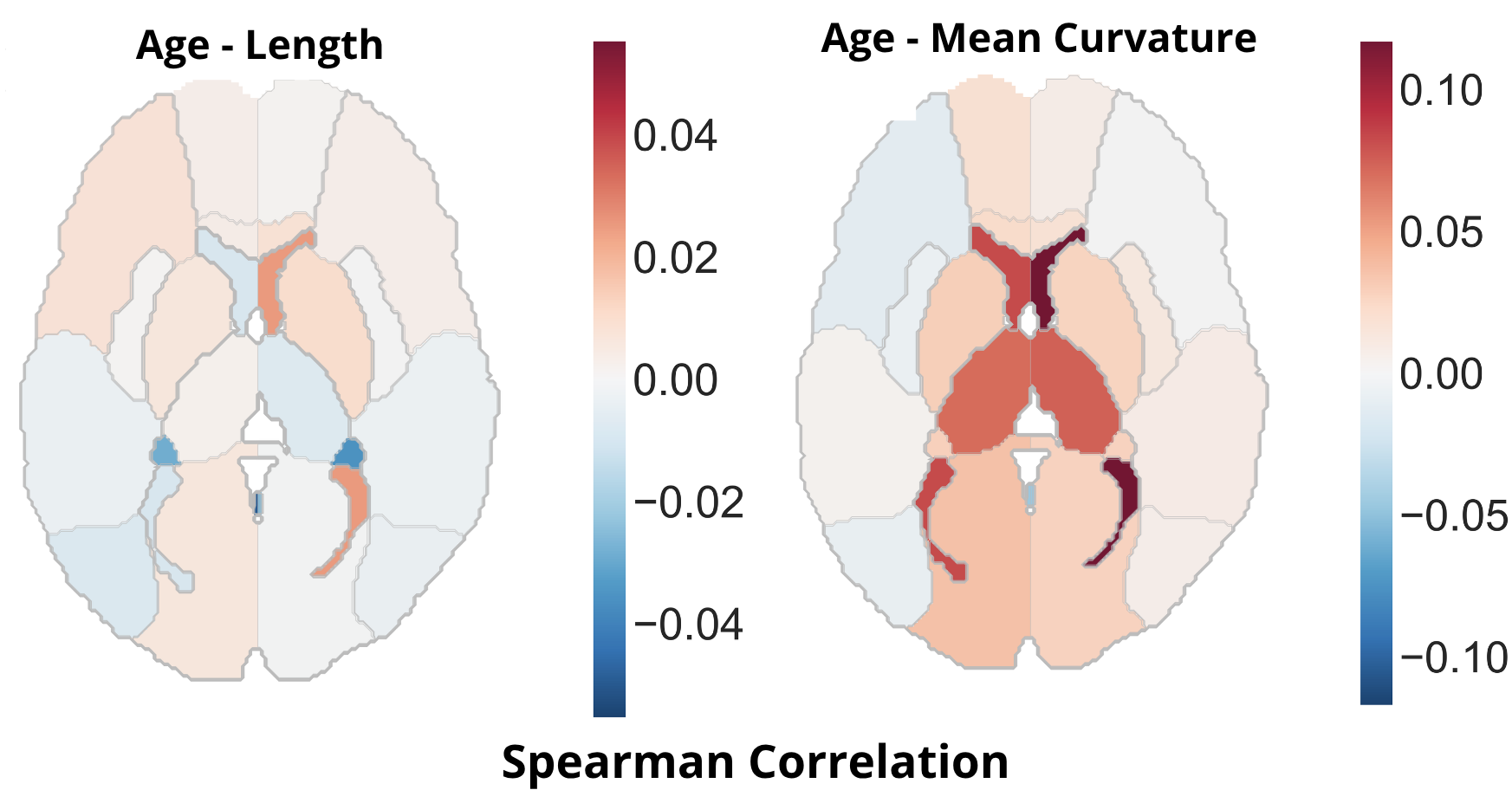}
    \caption{\textbf{Region-specific age effects.} Vessel length (left) and curvature (right) correlations across 30 brain arterial regions.}
    \label{fig:regional_analysis}
\end{figure}

\vspace{4pt}
\noindent\textbf{Results.}
Initial analysis revealed substantial site effects, particularly in curvature, total length, and volume ($\omega^2 = 0.723$, $0.406$, $0.289$), with IOP showing markedly different vessel distributions than Guys and HH (Figure~\ref{fig:site_effects}), likely reflecting scanner/acquisition variations~\cite{Mouches2019atlas}, though age imbalances may have contributed. Excluding IOP and focusing on the two homogeneous sites ($n=488$) strengthened age effects: total vessel length effect size increased from $\eta^2=0.095$ to $0.274$ (17.9 percentage points), suggesting site heterogeneity attenuates biological associations.
Analyzing age-related changes, negative correlations emerged between age and structural metrics (Fig.~\ref{fig:age_effects}). Total length, bifurcation count, and fractal dimension showed the sharpest declines ($r \approx -0.55$, $p < 0.001$), with total length exhibiting a ~20\% reduction across the lifespan (Fig.~\ref{fig:age_site_comparison}). Decreases in volume ($r \approx -0.38$), loop number ($r \approx -0.50$), and bifurcation density ($r \approx -0.35$) reflect network simplification and a loss of branching complexity exceeding vessel shortening. Conversely, curvature metrics ($r \approx +0.20$) and lacunarity ($r \approx +0.50$) increased with age, suggesting the remaining structure becomes more sparse and tortuous.
Quartile-based age analysis revealed large effect sizes for total length ($\eta^2 = 0.274$), bifurcations ($\eta^2 = 0.276$), volume ($\eta^2 = 0.133$), lacunarity ($\eta^2 = 0.235$), fractal dimension ($\eta^2 = 0.220$), and bifurcation density ($\eta^2 = 0.132$) (all $p < 0.001$), with a more modest association for mean curvature ($\eta^2 = 0.036$), confirming these features effectively capture structural vascular aging.
Regional analysis (Fig.~\ref{fig:regional_analysis}) revealed differences in correlation with age ($|r| \leq 0.30$) across cortical territories, with declines in length and fractal dimension with increased lacunarity, and deep nuclear regions exhibiting the opposite behavior.
\begin{figure}
    \centering
    \includegraphics[width=0.78\linewidth]{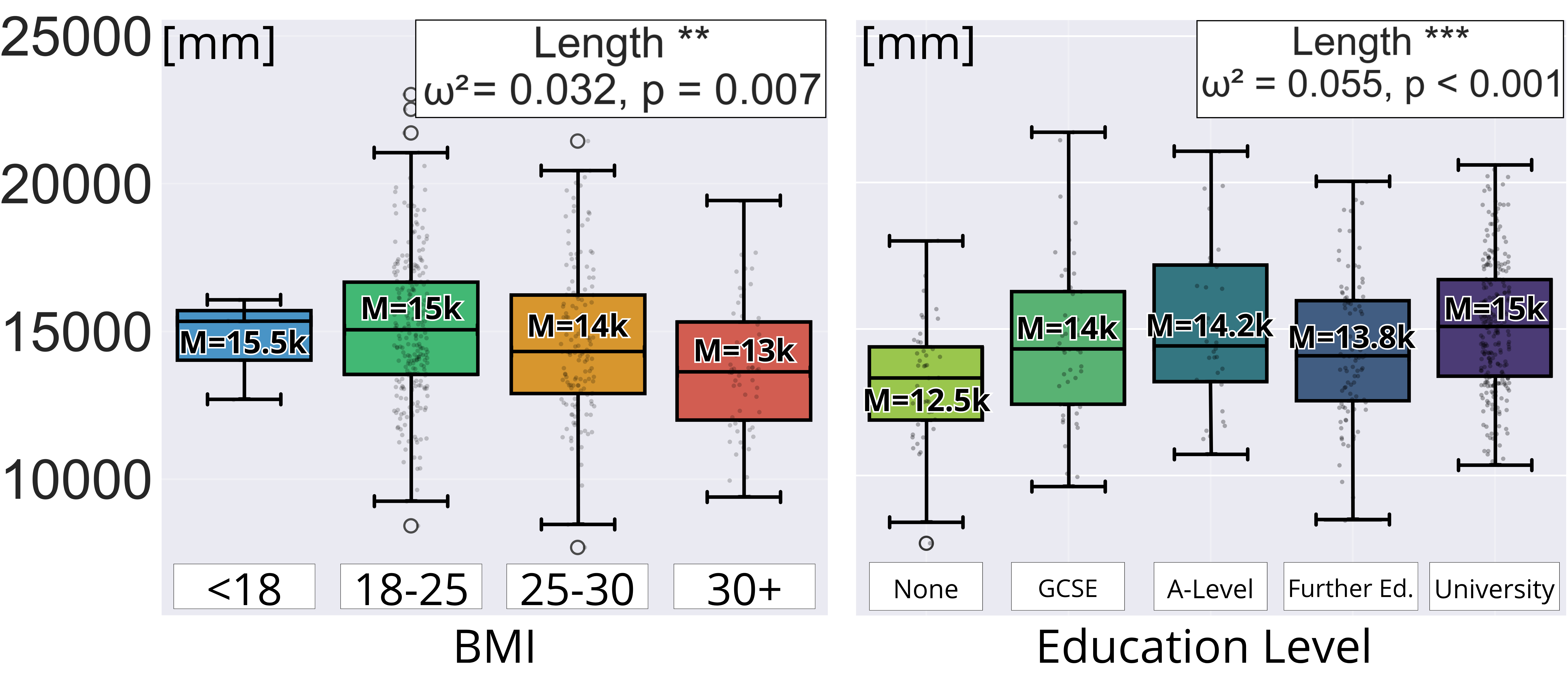}
    \caption{\textbf{Demographic effects on vessel length.} BMI (left) and education (right) gradients.}
    \label{fig:bmi_ed}
\end{figure}
Regarding anthropometry, normal-weight individuals ($n=252$) had longer vessels than overweight ($n=146$) or obese ($n=55$) groups ($\eta^2 = 0.060$, $p < 0.001$; Fig.~\ref{fig:bmi_ed}, left). Higher BMI was associated with structural simplification, shown by reduced bifurcation and node counts ($\eta^2 \approx 0.060$), increased curvature ($\eta^2 = 0.026$), and lacunarity ($\eta^2 = 0.059$). The underweight group ($n=3$) showed a similar pattern, though small sample size and age confounding limit the generalizability of the observations.
Height showed a significant positive association with vascular volume ($\eta^2 = 0.037, p = 0.007$). This is consistent with allometric scaling, where larger body size requires increased cerebral blood volume to perfuse a proportionally larger brain mass, without necessarily altering network complexity or density.
Post-IOP removal, females ($n=265$) exhibited higher fractal dimension ($d = 0.43$) and bifurcation density ($d = 0.31$) than males ($n=223$), suggesting more compact, complex networks that mirror height-related allometric scaling ($d = 0.39$ lacunarity in males).
Education analysis was prioritized over occupation to minimize age confounding. Results revealed a clear stepwise gradient, with higher qualification levels associated with increased vessel length ($\omega^2 = 0.077$, $p < 0.001$, $r=0.247$; Fig.~\ref{fig:bmi_ed}, right), increased fractal dimension ($\omega^2 = 0.045$) and reduced lacunarity ($\omega^2 = 0.043$, suggesting a more complex, spatially homogeneous organization consistent with cognitive reserve extending to cerebrovascular architecture~\cite{Stern2020cognitive}.

\section{Discussion and Conclusion}
We introduced \verb|CaravelMetrics|, an automated framework for cerebrovascular characterization that extracts 15 morphometric, topological, fractal, and geometric features directly from binary vessel masks, making it compatible with diverse neurovascular imaging modalities. Atlas integration enables both whole-brain and territory-specific analyses, enabling multiscale assessment of cerebrovascular organization.
Applied to 570 IXI subjects, \verb|CaravelMetrics| revealed age-related reductions in vessel length, volume, bifurcation count, and fractal dimension, alongside increased tortuosity and lacunarity, as well as education- and sex-associated network differences, consistent with literature regarding length reductions and increased tortuosity in healthy aging~\cite{Bullitt2010aging,ophelders2024anatomical}.
Limitations include sensitivity to mask quality and skeletonization. Future work should evaluate robustness across diverse protocols and pathological cohorts to validate these findings beyond a single healthy population.
\paragraph*{Compliance with ethical standards.}
\label{sec:ethics}
This study uses the publicly available datasets IXI and VesselVerse. No additional ethical approval was required.
\paragraph*{Acknowledgments.}
\label{sec:acknowledgments}
This work is co-funded by the European Union (ERC CARAVEL 101171357) and FLI (ANR-11-INBS-0006). LSC is supported by TwinsUK Imaging: A Resource for Ageing Research (Chronic Disease Research Foundation); MM and SO by King's Health Partners Digital Health Hub and the EPSRC. We acknowledge computational resources from GENCI-IDRIS on Jean Zay A100 partition (2025-AD011016543). We thank Bogdan Ion and Duccio Lalli for assistance with graph representation. 
Authors have no competing interests to declare. Views and opinions expressed are those of the authors and do not necessarily reflect those of the EU or the ERC. Neither the EU nor the granting authority can be held responsible for them.

\bibliographystyle{IEEEbib}
\bibliography{CaravelMetrics}

\end{document}